\documentclass[11pt,a4paper]{article}
\usepackage[hyperref]{acl2019}
\usepackage{times}
\usepackage{latexsym}
\usepackage{graphicx}
\usepackage{arydshln}
\usepackage{verbatim}
\usepackage{wrapfig}

\usepackage{url}
\usepackage{multirow}
\aclfinalcopy 
\def\aclpaperid{21} 

\title{Testing the Generalization Power of Neural Network Models Across NLI Benchmarks}

\author{Aarne Talman \\
   Department of Digital Humanities\\University of Helsinki \\
   Basement AI\\
  {\normalsize\tt aarne.talman@helsinki.fi} \And
  \bf Stergios Chatzikyriakidis\\
  Department of Philosophy, Linguistics and \\  Theory of Science, University of Gothenburg \\ 
      Basement AI\\
  {\normalsize\tt stergios.chatzikyriakidis@gu.se} \\}

\date{}

\begin{document}
\maketitle

\begin{abstract}
Neural network models have been very successful in natural language inference,
with the best models reaching 90\% accuracy in some benchmarks. However, the
success of these models turns out to be largely benchmark specific. We show that
models trained on a natural language inference dataset drawn from one benchmark
fail to perform well in others, even if the notion of inference assumed in these
benchmarks is the same or similar. We train six high performing neural network
models on different datasets and show that each one of these has problems of
generalizing when we replace the original test set with a test set taken from another
corpus designed for the same task. In light of these results, we argue that most
of the current neural network models are not able to generalize well in the task
of natural language inference. We find that using large pre-trained language models helps with transfer learning when the datasets are similar enough. Our results also highlight that the current NLI
datasets do not cover the different nuances of inference extensively enough. 
\end{abstract}

\section{Introduction}
Natural Language Inference (NLI) has attracted considerable interest in the NLP
community and, recently, a large number of neural network-based systems have
been proposed to deal with the task. One can attempt a rough categorization of
these systems into: a) sentence encoding systems, and b) other neural network
systems. Both of them have been very successful, with the state of the art on
the SNLI and MultiNLI datasets being 90.4\%, which is our baseline with BERT \cite{bert},
and 86.7\% \cite{bert} respectively. However, a big question with respect to
these systems is their ability to generalize outside the specific datasets they
are trained and tested on. Recently, \citet{breakingNLI} have shown that
state-of-the-art NLI systems break considerably easily when, instead of tested
on the original SNLI test set, they are tested on a test set which is
constructed by taking premises from the training set and creating several
hypotheses from them by changing at most one word within the premise. The
results show a very significant drop in accuracy for three of the four systems.
The system that was more difficult to break and had the least loss in accuracy
was the system by \citet{kim} which utilizes external knowledge taken from
WordNet \cite{Miller:1995}.

In this paper we show that NLI systems that have been very successful in
specific NLI benchmarks, fail to generalize when trained on a specific NLI
dataset and then these trained models are tested across test sets taken from
different NLI benchmarks. The results we get are in line with
\citet{breakingNLI}, showing that the generalization capability of the
individual NLI systems is very limited, but, what is more, they further show the
only system that was less prone to breaking in \citet{breakingNLI}, breaks too
in the experiments we have conducted.

We train six different state-of-the-art models on three different NLI datasets
and test these trained models on an NLI test set taken from another dataset
designed for the same NLI task, namely for the task to identify for sentence
pairs in the dataset if one sentence entails the other one, if they are in
contradiction with each other or if they are neutral with respect to inferential
relationship.

One would expect that if a model learns to correctly identify inferential
relationships in one dataset, then it would also be able to do so in another
dataset designed for the same task. Furthermore, two of the datasets, SNLI
\cite{snli} and MultiNLI \cite{multinli}, have been constructed using the
same crowdsourcing approach and annotation instructions \cite{multinli}, leading
to datasets with the same or at least very similar definition of entailment. It
is therefore reasonable to expect that transfer learning between these datasets
is possible. As SICK \cite{sick} dataset has been machine-constructed, a bigger
difference in performance is expected.

In this paper we show that, contrary to our expectations, most models fail to
generalize across the different datasets. However, our experiments also show
that BERT \citep{bert} performs much better than the other models in experiments
between SNLI and MultiNLI. Nevertheless, even BERT fails when testing on SICK.
In addition to the negative results, our experiments further highlight the power
of pre-trained language models, like BERT, in NLI.

The negative results of this paper are significant for the NLP research
community as well as to NLP practice as we would like our best models to not
only to be able to perform well in a specific benchmark dataset, but rather
capture the more general phenomenon this dataset is designed for. The main
contribution of this paper is that it shows that most of the best performing
neural network models for NLI fail in this regard. The second, and equally
important, contribution is that our results highlight that the current NLI
datasets do not capture the nuances of NLI extensively enough.

\section{Related Work}
The ability of NLI systems to generalize and related skepticism has been raised
in a number of recent papers.  \citet{breakingNLI}  show that the generalization
capabilities of state-of-the-art NLI systems, in cases where some kind of
external lexical knowledge is needed, drops dramatically when the SNLI test set
is replaced by a test set where the premise and the hypothesis are otherwise
identical except for at most one word. The results show a very significant drop
in accuracy. \citet{kang2018} recognize the generalization problem that comes
with training on datasets like SNLI, which tend to be \textit{homogeneous} and
with little linguistic variation. In this context, they propose to better train
NLI models by making use of adversarial examples.

Multiple papers have reported hidden bias and annotation artifacts in the
popular NLI datasets SNLI and MultiNLI allowing classification based on the
hypothesis sentences alone \citep{TSUCHIYA18, gururangan:2018, poliak2018}.

\citet{wang2018simply} evaluate the robustness of NLI models using datasets
where label preserving swapping operations have been applied, reporting
significant performance drops compared to the results with the original dataset.
In these experiments, like in the BreakingNLI experiment, the systems that seem
to be performing the better, i.e. less prone to breaking, are the ones where
some kind of external knowledge is used by the model (KIM by \citealp{kim} is one of
those systems).

On a theoretical and methodological level, there is discussion on the nature of
various NLI datasets, as well as the definition of what counts as NLI and what
does not. For example, \citet{chatzikyriakidis2017,bernardy:2019} present an
overview of the most standard datasets for NLI and show that the definitions of
inference in each of them are actually quite different, capturing only fragments
of what seems to be a more general phenomenon.

 \citet{snli} show that a simple LSTM model trained on the SNLI data fails when
 tested on SICK. However, their experiment is limited to this single
 architecture and dataset pair. \citet{multinli} show that different models that
 perform well on SNLI have lower accuracy on MultiNLI. However in their
 experiments they did not systematically test transfer learning between the two datasets, but
 instead used separate systems where the training and test data were drawn from
 the same corpora.\footnote{To be more precise, \citet{multinli} tested some transfer across datasets, but only between MultiNLI and SNLI.}
 
 \begin{table*}[t!]
\centering
        \begin{tabular}{l l | c c}
            \hline
            \bf Train data&\bf Test data & \bf Size of the training set &  \bf Size of the test set\\
            \hline
            \bf SNLI &\bf SNLI & 550,152 & 10,000\\
            SNLI & MultiNLI & 550,152 & 20,000\\
            SNLI  & SICK & 550,152 & 9,840\\
            \hdashline
            \bf MultiNLI &\bf MultiNLI & 392,702 & 20,000\\
            MultiNLI & SNLI &  392,702 & 10,000\\
            MultiNLI & SICK & 392,702 & 9,840\\
            \hdashline
            \bf SNLI + MultiNLI &\bf SNLI & 942,854 & 10,000\\
            SNLI + MultiNLI & SICK & 942,854 & 9,840\\
            \hline
        \end{tabular}
    \caption{\label{table:data} Dataset combinations used in the experiments. The rows in bold are baseline experiments, where the test data comes from the same benchmark as the training and development data.}
\end{table*}

\section{Experimental Setup}
In this section we describe the datasets and model architectures included in the experiments.

\subsection{Data}
We chose three different datasets for the experiments: SNLI, MultiNLI and SICK.
All of them have been designed for NLI involving three-way classification with
the labels \emph{entailment}, \emph{neutral} and \emph{contradiction}. We did
not include any datasets with two-way classification, e.g. SciTail
\cite{scitail}.  As SICK is a relatively small dataset with approximately only
10k sentence pairs, we did not use it as training data in any experiment. We
also trained the models with a combined SNLI + MultiNLI training set.

For all the datasets we report the baseline performance where the training and
test data are drawn from the same corpus. We then take these trained models and
test them on a test set taken from another NLI corpus. For the case where the models are trained with SNLI + MultiNLI we report the baseline using the SNLI test data.\footnote{Here we could as well have selected MultiNLI test data. However the selection does not impact our findings.} All the experimental
combinations are listed in Table \ref{table:data}. Examples from the selected
datasets are provided in Table \ref{table:examples}. To be more precise, we vary three things: training dataset, model and testing dataset. We should qualify this though, since the three datasets we look at,  can also be grouped by text domain/genre and type of data collection, with MultiNLI and SNLI using the same data collection style, and SNLI and SICK using roughly the same domain/genre. Hopefully,  our set up will let us determine which of these factors matters the most.

We describe the source datasets in more detail below.

\subsubsection*{SNLI}
The Stanford Natural Language Inference (SNLI) corpus \cite{snli} is a dataset
of 570k human-written sentence pairs manually labeled with the labels
entailment, contradiction, and neutral. The source for the premise sentences in
SNLI were image captions taken from the Flickr30k corpus \cite{Flickr:TACL229}.

\subsubsection*{MultiNLI}
The Multi-Genre Natural Language Inference (MultiNLI) corpus \cite{multinli}
consisting of 433k human-written sentence pairs labeled with entailment,
contradiction and neutral. MultiNLI contains sentence pairs from ten distinct
genres of both written and spoken English. Only five genres are included in the
training set. The development and test sets have been divided into matched and
mismatched, where the former includes only sentences from the same genres as the
training data, and the latter includes sentences from the remaining genres not
present in the training data.

We used the matched development set (MultiNLI-m) for the
experiments.\footnote{Here the choice between MultiNLI matched and mismatched
does not make a difference to our experimental setup.} The MultiNLI dataset was
annotated using very similar instructions as for the SNLI dataset.\footnote{The reason we are not saying ``exactly the same" is because in some cases, as the authors  report, ``the prompts that surround each premise sentence during hypothesis collection are slightly tailored to fit the genre of that premise sentence''.} Therefore we can
assume that the definitions of entailment, contradiction and neutral is the same
in these two datasets.

\subsubsection*{SICK}
SICK \citep{sick} is a dataset that was  originally constructed to test
compositional distributional semantics (DS) models. The dataset contains 9,840
examples pertaining to logical inference (negation, conjunction, disjunction,
apposition, relative clauses, etc.). The dataset was automatically constructed
taking pairs of sentences from a random subset of the 8K ImageFlickr data set
\cite{Flickr:TACL229} and the SemEval 2012 STS MSRVideo Description dataset
\cite{semeval2012-task6}.

\begin{table*}[ht!]
\begin{tabular}{l l}
\hline
 &\bf entailment \\
\hline
SICK & \sl A person, who is riding a bike, is wearing gear which is black\\
 & \sl A biker is wearing gear which is black \\
 \hdashline
 SNLI & \sl A young family enjoys feeling ocean waves lap at their feet.\\
  & \sl A family is at the beach. \\
   \hdashline
MultiNLI & \sl Kal tangled both of Adrin's arms, keeping the blades far away.\\
 & \sl Adrin's arms were tangled, keeping his blades away from Kal. \\
\hline
 &\bf contradiction\\
\hline
SICK & \sl There is no man wearing a black helmet and pushing a bicycle\\
 & \sl One man is wearing a black helmet and pushing a bicycle \\
  \hdashline
 SNLI & \sl A man with a tattoo on his arm staring to the side with vehicles and buildings behind him.\\
 & \sl A man with no tattoos is getting a massage. \\
 \hdashline
MultiNLI & \sl Also in Eustace Street is an information office and a cultural center for children, The Ark .\\
 & \sl The Ark, a cultural center for kids, is located in Joyce Street. \\
\hline
&\bf neutral \\
\hline
SICK & \sl A little girl in a green coat and a boy holding a red sled are walking in the snow\\
 & \sl A child is wearing a coat and is carrying a red sled near a child in a green and black coat\\
 \hdashline
 SNLI & \sl An old man with a package poses in front of an advertisement.\\
 & \sl A man poses in front of an ad for beer. \\
 \hdashline
 MultiNLI & \sl Enthusiasm for Disney's Broadway production of The Lion King dwindles.\\
 & \sl The broadway production of The Lion King was amazing, but audiences are getting bored. \\
\hline
\end{tabular}
\caption{\label{table:examples} Example sentence pairs from the three datasets.}
\end{table*}
 
\subsection{Model and Training Details}
We perform experiments with six high-performing models covering the sentence
encoding models, cross-sentence attention models as well as fine-tuned
pre-trained language models.

For sentence encoding models, we chose a simple one-layer bidirectional LSTM
with max pooling (BiLSTM-max) with the hidden size of 600D per direction, used
e.g. in InferSent \citep{infersent}, and HBMP \cite{talman2018hbmp}. For the
other models, we have chosen ESIM \citep{Chen}, which includes cross-sentence
attention, and KIM \cite{kim}, which has cross-sentence attention and utilizes
external knowledge. We also selected two model involving a pre-trained language
model, namely ESIM + ELMo \citep{Peters:2018} and BERT \citep{bert}. KIM is
particularly interesting in this context as it performed significantly better
than other models in the Breaking NLI experiment conducted by
\citet{breakingNLI}. The success of pre-trained language models in multiple NLP
tasks make ESIM + ELMo and BERT interesting additions to this experiment. Table
\ref{table:models} lists the different models used in the experiments.

For BiLSTM-max we used the Adam optimizer \cite{KingmaB14Adam}, a learning rate
of 5e-4 and batch size of 64. The learning rate was decreased by the factor of
0.2 after each epoch if the model did not improve. Dropout of 0.1 was used
between the layers of the multi-layer perceptron classifier, except before the
last layer.The BiLSTM-max models were initialized with pre-trained GloVe 840B
word embeddings of size 300 dimensions \cite{pennington2014glove}, which were
fine-tuned during training. Our BiLSMT-max model was implemented in PyTorch.

For HBMP, ESIM, KIM and BERT we used the original implementations with the
default settings and hyperparameter values as described in
\citet{talman2018hbmp}, \citet{Chen}, \citet{kim} and \citet{bert} respectively.
For BERT we used the uncased 768-dimensional model (BERT-base). For ESIM + ELMo
we used the AllenNLP \citep{Gardner2017AllenNLP} PyTorch implementation with the
default settings and hyperparameter values.

 \begin{table*}[ht!]
\centering
        \begin{tabular}{l l}
            \hline
            \bf Model & {\bf Model type}\\
            \hline
                BiLSTM-max \citep{infersent} & Sentence encoding \\
                HBMP \cite{talman2018hbmp} & Sentence encoding\\
                \hdashline
                ESIM \cite{Chen} & Cross-sentence attention\\
                KIM \cite{kim} & Cross-sentence attention\\
                \hdashline
                ESIM + ELMo \cite{Peters:2018} & Pre-trained language model\\
                BERT-base \cite{bert} & Cross-sentence attention + pre-trained language model\\
            \hline
        \end{tabular}
    \caption{\label{table:models} Model architectures used in the experiments. }
\end{table*}

\section{Experimental Results}
Table \ref{table:neg_encoding} contains all the experimental results.

Our experiments show that, while all of the six models perform well when the
test set is drawn from the same corpus as the training and development set,
accuracy is significantly lower when we test these trained models on a test set
drawn from a separate NLI corpus, the average difference in accuracy being 24.9
points across all experiments.

Accuracy drops the most when a model is tested on SICK. The difference in this
case is between 19.0-29.0 points when trained on MultiNLI, between 31.6-33.7
points when trained on SNLI and between 31.1-33.0 when trained on SNLI +
MultiNLI. This was expected, as the method of constructing the sentence pairs
was different, and hence there is too much difference in the kind of sentence
pairs included in the training and test sets for transfer learning to work.
However, the drop was more dramatic than expected.

The most surprising result was that the accuracy of all models drops
significantly even when the models were trained on MultiNLI and tested on SNLI
(3.6-11.1 points). This is surprising as both of these datasets have been
constructed with a similar data collection method using the same definition of
entailment, contradiction and neutral. The sentences included in SNLI are also
much simpler compared to those in MultiNLI, as they are taken from the Flickr
image captions. This might also explain why the difference in accuracy for all
of the six models is lowest when the models are trained on MultiNLI and tested
on SNLI. It is also very surprising that the model with the biggest difference
in accuracy was ESIM + ELMo which includes a pre-trained ELMo language model.
BERT performed significantly better than the other models in this experiment
having an accuracy of 80.4\% and only 3.6 point difference in accuracy.

The poor performance of most of the models with the MultiNLI-SNLI dataset pair
is also very surprising given that neural network models do not seem to suffer a
lot from introduction of new genres to the test set which were not included in
the training set, as can be seen from the small difference in test accuracies
for the matched and mismatched test sets (see e.g \citealp{multinli}). In a sense
SNLI could be seen as a separate genre not included in MultiNLI. This raises the
question if the SNLI and MultiNLI have e.g. different kinds of annotation
artifacts, which makes transfer learning between these datasets more difficult.

All the models, except BERT, perform almost equally poorly across all the
experiments. Both BiLSTM-max and HBMP have an average drop in accuracy of 24.4
points, while the average for KIM is 25.5 and for ESIM + ELMo 25.6. ESIM has the
highest average difference of 27.0 points. In contrast to the findings of
\citet{breakingNLI}, utilizing external knowledge did not improve the model's
generalization capability, as KIM performed equally poorly across all dataset
combinations.

Also including a pretrained ELMo language model did not improve the results
significantly. The overall performance of BERT was significantly better than the
other models, having the lowest average difference in accuracy of 22.5 points.
Our baselines for SNLI (90.4\%) and SNLI + MultiNLI (90.6\%) outperform the
previous state-of-the-art accuracy for SNLI (90.1\%) by \citet{kim:2018}.

\begin{table*}[ht!]
\small
\centering
        \begin{tabular}{l l |c c| l}
            \hline
            \bf Train data\hspace{1,5cm}  &\bf Test data \hspace{0,5cm} &\bf Test accuracy  &\bf \hspace{0,5cm} $\Delta$\hspace{1cm}  &\bf Model \\
            \hline
            \bf SNLI&\bf SNLI &\bf 86.1 &\bf &{\bf BiLSTM-max} (our baseline)\\
            \bf SNLI &\bf SNLI &\bf 86.6 &\bf &{\bf HBMP} \cite{talman2018hbmp}\\
            \bf SNLI &\bf SNLI &\bf 88.0 &\bf &{\bf ESIM} \cite{Chen}\\
            \bf SNLI &\bf SNLI &\bf 88.6 &\bf &{\bf KIM} \cite{kim}\\
            \bf SNLI &\bf SNLI &\bf 88.6 &\bf &{\bf ESIM + ELMo} \cite{Peters:2018}\\
            \bf SNLI &\bf SNLI &\bf \underline{90.4} &\bf &{\bf BERT-base} \cite{bert}\\
            \hdashline
            SNLI & MultiNLI-m & 55.7\textsuperscript{*} & -30.4 & BiLSTM-max\\
            SNLI & MultiNLI-m & 56.3\textsuperscript{*} & -30.3 & HBMP\\
            SNLI & MultiNLI-m & 59.2\textsuperscript{*} & -28.8 & ESIM \\
            SNLI& MultiNLI-m & 61.7\textsuperscript{*} & -26.9 & KIM \\
            SNLI & MultiNLI-m & 64.2\textsuperscript{*} & -24.4 & ESIM + ELMo \\
            SNLI & MultiNLI-m &  \underline{75.5}\textsuperscript{*} & \underline{-14.9} & BERT-base\\
            \hdashline
            SNLI& SICK & 54.5 & \underline{-31.6} & BiLSTM-max \\
            SNLI & SICK & 53.1 & -33.5 & HBMP \\
            SNLI & SICK & 54.3 & -33.7 & ESIM \\
            SNLI & SICK & 55.8 & -32.8 & KIM \\
            SNLI & SICK & 56.7 & -31.9 & ESIM + ELMo \\
            SNLI & SICK & \underline{56.9} & -33.5 & BERT-base \\
            \hline
            \bf MultiNLI &\bf MultiNLI-m &\bf 73.1\textsuperscript{*} & &\bf BiLSTM-max \\
            \bf MultiNLI &\bf MultiNLI-m &\bf 73.2\textsuperscript{*} & &\bf HBMP \\
            \bf MultiNLI &\bf MultiNLI-m &\bf 76.8\textsuperscript{*} & &\bf ESIM \\
            \bf MultiNLI &\bf MultiNLI-m &\bf 77.3\textsuperscript{*} & &\bf KIM \\
            \bf MultiNLI &\bf MultiNLI-m &\bf 80.2\textsuperscript{*} & &\bf ESIM + ELMo \\
            \bf MultiNLI &\bf MultiNLI-m &\bf \underline{84.0}\textsuperscript{*} & &\bf BERT-base \\
            \hdashline
            MultiNLI & SNLI & 63.8 & -9.3 & BiLSTM-max \\
            MultiNLI & SNLI & 65.3 & -7.9 & HBMP \\
            MultiNLI & SNLI & 66.4 & -10.4 & ESIM \\
            MultiNLI & SNLI & 68.5 & -8.8 & KIM \\
            MultiNLI & SNLI &  69.1 &  -11.1 & ESIM + ELMo\\
            MultiNLI & SNLI &  \underline{80.4} &  \underline{-3.6}  & BERT-base\\
            \hdashline
            MultiNLI & SICK & 54.1 & \underline{-19.0} & BiLSTM-max \\
            MultiNLI & SICK & 54.1 & -19.1 & HBMP \\
            MultiNLI & SICK & 47.9 & -28.9 & ESIM \\
            MultiNLI & SICK & 50.9 & -26.4 & KIM \\
            MultiNLI & SICK & 51.4 & -28.8 & ESIM + ELMo\\
            MultiNLI & SICK & \underline{55.0} & -29.0 & BERT-base\\
            \hline
            \bf SNLI + MultiNLI &\bf SNLI &\bf 86.1 & &\bf BiLSTM-max \\
            \bf SNLI + MultiNLI &\bf SNLI &\bf 86.1 & &\bf HBMP \\
            \bf SNLI + MultiNLI &\bf SNLI &\bf 87.5 & &\bf ESIM \\
            \bf SNLI + MultiNLI &\bf SNLI &\bf 86.2 & &\bf KIM \\
            \bf SNLI + MultiNLI &\bf SNLI &\bf 88.8 & &\bf ESIM + ELMo\\
            \bf SNLI + MultiNLI &\bf SNLI &\bf 90.6 & &\bf BERT-base\\
            \hdashline
            SNLI + MultiNLI & SICK & 54.5 & -31.6 & BiLSTM-max \\
            SNLI + MultiNLI & SICK & 55.0 & \underline{-31.1} & HBMP \\
            SNLI + MultiNLI & SICK & 54.5 & -33.0 & ESIM \\
            SNLI + MultiNLI & SICK & 54.6 & -31.6 & KIM \\ 
            SNLI + MultiNLI & SICK & 57.1 & -31.7 & ESIM + ELMo\\
            SNLI + MultiNLI & SICK & \underline{59.1} & -31.5 & BERT-base\\
            \hline
        \end{tabular}
    \caption{\label{table:neg_encoding} Test accuracies (\%). For the baseline results (highlighted in bold) the training data and test data have been drawn from the same benchmark corpus. $\Delta$ is the difference between the test accuracy and the baseline accuracy for the same training set. Results marked with \textsuperscript{*} are for the development set, as no annotated test set is openly available. Best scores with respect to accuracy and difference in accuracy are underlined.}
\end{table*}

To understand better the types of errors made by neural network models in NLI we looked at some example failure-pairs for selected models.\footnote{More thorough error analysis of each of the models and set-up is out of scope of this work but we intend to address these in our future research.} Tables \ref{table:error_examples_bert} and \ref{table:error_examples_hbmp} contain some randomly selected failure-pairs for two models: BERT and HBMP, and for three set-ups: SNLI$\rightarrow$SICK, SNLI$\rightarrow$MultiNLI and MultiNLI$\rightarrow$SICK. We chose BERT as the current the state of the art NLI model. HBMP was selected as a high performing model in the sentence encoding model type. Although the listed sentence pairs represent just a small sample of the errors made by these models, they do include some interesting examples. First, it seems that SICK has a more narrow notion of contradiction -- corresponding more to logical contradiction -- compared to the contradiction in SNLI and MultiNLI, where especially in SNLI the sentences are contradictory if they describe a different state of affairs. This is evident in the sentence pair: \emph{A young child is running outside over the fallen leaves} and \emph{A young child is lying down on a gravel road that is covered with dead leaves}, which is predicted by BERT to be \emph{contradiction} although the gold label is \emph{neutral}. Another interesting example is the sentence pair: \emph{A boat pear with people boarding and disembarking some boats.} and \emph{people are boarding and disembarking some boats}, which is incorrectly predicted by BERT to be \emph{contradiction} although it has been labeled as \emph{entailment}. Here the two sentences describe the same event from different points of view: the first one describing a boat pear with some people on it and the second one describing the people directly. Interestingly the added information about the boat pear seems to confuse the model.

\section{Discussion and Conclusion}
In this paper we have shown that neural network models for NLI fail to
generalize across different NLI benchmarks. We experimented with six
state-of-the-art models covering sentence encoding approaches, cross-sentence
attention models and pre-trained and fine-tuned language models. For all the
systems, the accuracy drops between 3.6-33.7 points (the average drop being 24.9
points), when testing with a test set drawn from a separate corpus from that of
the training data, as compared to when the test and training data are splits
from the same corpus. Our findings, together with the previous negative
findings, indicate that the state-of-the-art models fail to capture the
semantics of NLI in a way that will enable them to generalize across different
NLI situations.

The results highlight two issues to be taken into consideration: a) using
datasets involving a fraction of what NLI is, will fail when tested in datasets
that are testing for a slightly different definition of inference. This is
evident when we move from the SNLI to the SICK dataset. b) NLI is to some extent
genre/context dependent. Training on SNLI and testing on MultiNLI gives worse
results than vice versa. This is particularly evident in the case of BERT. These
results highlight that training on multiple genres helps.  However, this help is
still not enough given that, even in the case of training on MultiNLI (multi
genre) and training  on SNLI  (single genre  and same  definition of inference
with MultiNLI), accuracy drops significantly.

We also found that involving a large pre-trained language model helps with
transfer learning when the datasets are similar enough, as is the case with SNLI
and MultiNLI. Our results further corroborate the power of pre-trained and
fine-tuned language models like BERT in NLI. However, not even BERT is able to
generalize from SNLI and MultiNLI to SICK, possibly due to the difference
between what kind of inference relations are contained in these datasets.

Our findings motivate us to look for novel neural network architectures and
approaches that better capture the semantics on natural language inference
beyond individual datasets. However, there seems to be a need to start with
better constructed datasets, i.e. datasets that will not only capture fractions
of what NLI is in reality. Better NLI systems need to be able to be more
versatile on the types of inference they can recognize. Otherwise, we would be
stuck with systems that can cover only some aspects of NLI. 
On a theoretical level, and in connection to the previous point, we need a
better understanding of the range of phenomena NLI must be able to cover and
focus our future endeavours for dataset construction towards this direction. In
order to do this a more systematic study is needed on the different kinds of
entailment relations NLI datasets need to include. Our future work will include a more systematic and broad-coverage analysis of the types of errors the models make and in what kinds of sentence-pairs they make successful predictions.

\begin{table*}[ht!]
\begin{tabular}{l l}
\hline
\multicolumn{2}{l}{\bf BERT: SNLI $\rightarrow$ SICK }\\
\hline
\multicolumn{2}{l}{A young child is running outside over the fallen leaves} \\
\multicolumn{2}{l}{A young child is lying down on a gravel road that is covered with dead leaves }\\
\hdashline
{\bf Label:} neutral & {\bf Prediction:} contradiction \\
\hline
\multicolumn{2}{l}{The man is being knocked off of a horse}	\\
\multicolumn{2}{l}{Someone is falling off a horse} \\
\hdashline
{\bf Label:} entailment & {\bf Prediction:} contradiction \\
\hline
\multicolumn{2}{l}{There is no one typing} \\
\multicolumn{2}{l}{Someone is typing on a keyboard} \\
\hdashline
{\bf Label:} contradiction & {\bf Prediction:} neutral \\
\hline
\multicolumn{2}{l}{The man in the purple hat is operating a camera that makes videos} \\
\multicolumn{2}{l}{There is no man with a camera studying the subject} \\
\hdashline
{\bf Label:} neutral & {\bf Prediction:} contradiction \\
\hline
\multicolumn{2}{l}{A woman is taking off a cloak, which is very large, and revealing an extravagant dress} \\
\multicolumn{2}{l}{A woman is putting on a cloak, which is very large, and concealing an extravagant dress}\\
\hdashline
{\bf Label:} contradiction & {\bf Prediction:} neutral \\
\hline
~\\
\hline
\multicolumn{2}{l}{\bf BERT: MultiNLI $\rightarrow$ SICK }\\
\hline
\multicolumn{2}{l}{A cowboy is riding a horse and cornering a barrel} \\
\multicolumn{2}{l}{A cowgirl is riding a horse and corners a barrel}\\
\hdashline
{\bf Label:} neutral & {\bf Prediction:} contradiction \\
\hline
\multicolumn{2}{l}{A tan dog is jumping up and catching a tennis ball}	\\
\multicolumn{2}{l}{A dog with a tan coat is jumping up and catching a tennis ball} \\
\hdashline
{\bf Label:} entailment & {\bf Prediction:} neutral \\
\hline
\multicolumn{2}{l}{The bunch of men are playing rugby on a muddy field} \\
\multicolumn{2}{l}{Some men are idling} \\
\hdashline
{\bf Label:} neutral & {\bf Prediction:} contradiction \\
\hline
\multicolumn{2}{l}{A blond child is going down a slide and throwing up his arms} \\
\multicolumn{2}{l}{A child with dark hair is going down a slide and throwing up his arms} \\
\hdashline
{\bf Label:} contradiction & {\bf Prediction:} entailment \\
\hline
\multicolumn{2}{l}{There is no person in bike gear standing steadily in front of the mountains} \\
\multicolumn{2}{l}{A group of people is equipped with protective gear}\\
\hdashline
{\bf Label:} neutral & {\bf Prediction:} contradiction \\
\hline
~\\
\hline
\multicolumn{2}{l}{\bf BERT: MultiNLI $\rightarrow$ SNLI}\\
\hline
\multicolumn{2}{l}{A woman in a white wedding dress is being dressed and fitted by two other women.} \\
\multicolumn{2}{l}{A woman is being fitted for the first time.}\\
\hdashline
{\bf Label:} neutral   & {\bf Prediction:} contradiction   \\
\hline
\multicolumn{2}{l}{A boat pear with people boarding and disembarking some boats.} \\
\multicolumn{2}{l}{people are boarding and disembarking some boats}\\
\hdashline
{\bf Label:} entailment   & {\bf Prediction:} contradiction   \\
\hline
\multicolumn{2}{l}{Several men at a bar watch a sports game on the television.	} \\
\multicolumn{2}{l}{The men are at a baseball game.}\\
\hdashline
{\bf Label:} contradiction   & {\bf Prediction:} entailment  \\
\hline
\multicolumn{2}{l}{A singer wearing a leather jacket performs on stage with dramatic lighting behind him.} \\
\multicolumn{2}{l}{a singer is on american idol}\\
\hdashline
{\bf Label:} neutral   &{\bf Prediction:} contradiction   \\
\hline
\multicolumn{2}{l}{A person rolls down a hill riding a wagon as another watches.} \\
\multicolumn{2}{l}{A person stares at an empty hill.}\\
\hdashline
{\bf Label:} contradiction   & {\bf Prediction:} neutral   \\
\hline
\end{tabular}
\caption{\label{table:error_examples_bert} Example failure-pairs for BERT.}
\end{table*}

\begin{table*}[ht!]
\begin{tabular}{l l}
\hline
\multicolumn{2}{l}{\bf HBMP: SNLI $\rightarrow$ SICK }\\
\hline
\multicolumn{2}{l}{a boy is sitting in a room and playing a piano by lamp light} \\
\multicolumn{2}{l}{a boy is playing a keyboard}\\
\hdashline
{\bf Label:} entailment  & {\bf Prediction:} contradiction  \\
\hline
\multicolumn{2}{l}{a woman is wearing ear protection and is firing a gun at an outdoor shooting range} \\
\multicolumn{2}{l}{a woman is shooting at target practices}\\
\hdashline
{\bf Label:} entailment  &{\bf Prediction:} neutral  \\
\hline
\multicolumn{2}{l}{a man in a purple hat isn't climbing a rocky wall with bare hands} \\
\multicolumn{2}{l}{a man in a purple hat is climbing a rocky wall with bare hands}\\
\hdashline
{\bf Label:} contradiction  & {\bf Prediction:} entailment  \\
\hline
\multicolumn{2}{l}{a cat is swinging on a fan} \\
\multicolumn{2}{l}{a cat is stuck on a moving ceiling fan}\\
\hdashline
{\bf Label:} neutral  & {\bf Prediction:} contradiction  \\
\hline
\multicolumn{2}{l}{a young boy is jumping and covering nearby wooden fence with grass} \\
\multicolumn{2}{l}{a young boy covered in grass is jumping near a wooden fence}\\
\hdashline
{\bf Label:} neutral  & {\bf Prediction:} entailment  \\
\hline
~\\
\hline
\multicolumn{2}{l}{\bf HBMP: MultiNLI $\rightarrow$ SICK }\\
\hline
\multicolumn{2}{l}{two dogs are walking slowly through a park} \\
\multicolumn{2}{l}{two dogs are running quickly through a park}\\
\hdashline
{\bf Label:} neutral  & {\bf Prediction:} entailment  \\
\hline
\multicolumn{2}{l}{the woman is playing a guitar which is electric} \\
\multicolumn{2}{l}{the woman is playing an electric guitar}\\
\hdashline
{\bf Label:} entailment  & {\bf Prediction:} neutral  \\
\hline
\multicolumn{2}{l}{there is no man squatting in brush and taking a photograph} \\
\multicolumn{2}{l}{a man is crouching and holding a camera}\\
\hdashline
{\bf Label:} neutral  & {\bf Prediction:} contradiction  \\
\hline
\multicolumn{2}{l}{the snowboarder is jumping off a snowy hill} \\
\multicolumn{2}{l}{a snowboarder is jumping off the snow}\\
\hdashline
{\bf Label:} neutral  & {\bf Prediction:} entailment  \\
\hline
\multicolumn{2}{l}{the boy is sitting near the blue ocean} \\
\multicolumn{2}{l}{the boy is wading through the blue ocean}\\
\hdashline
{\bf Label:} contradiction  & {\bf Prediction:} neutral  \\
\hline
~\\
\hline
\multicolumn{2}{l}{\bf HBMP: MultiNLI $\rightarrow$ SNLI }\\
\hline
\multicolumn{2}{l}{a man is holding a book standing in front of a chalkboard.} \\
\multicolumn{2}{l}{a person is in a classroom teaching.}\\
\hdashline
{\bf Label:} entailment  & {\bf Prediction:} contradiction  \\
\hline
\multicolumn{2}{l}{a woman with a pink purse walks down a crowded sidewalk.} \\
\multicolumn{2}{l}{a woman escapes a from a hostile enviroment}\\
\hdashline
{\bf Label:} neutral  & {\bf Prediction:} contradiction  \\
\hline
\multicolumn{2}{l}{a woman with a pink purse walks down a crowded sidewalk.} \\
\multicolumn{2}{l}{a woman escapes a from a hostile enviroment}\\
\hdashline
{\bf Label:} neutral  & {\bf Prediction:} contradiction  \\
\hline
\multicolumn{2}{l}{a person waterskiing in a river with a large wall in the background.} \\
\multicolumn{2}{l}{a dog waterskiing in a river with a large wall in the background.}\\
\hdashline
{\bf Label:} contradiction  & {\bf Prediction:} neutral  \\
\hline
\multicolumn{2}{l}{a man wearing a blue shirt and headphones around his neck raises his arm.} \\
\multicolumn{2}{l}{a man is raising his arm to get someones attention.}\\
\hdashline
{\bf Label:} neutral  & {\bf Prediction:} entailment  \\
\hline
\end{tabular}
\caption{\label{table:error_examples_hbmp} Example failure-pairs for HBMP.}
\end{table*}

\section*{Acknowledgments}
\vspace{1ex}
\noindent
\begin{minipage}{0.1\linewidth}
   \raisebox{-0.2\height}{\includegraphics[trim =32mm 55mm 30mm 5mm, clip, scale=0.2]{erc.ai}}
\end{minipage}
\hspace{0.01\linewidth}
\begin{minipage}{0.70\linewidth}
The first author is supported by the FoTran project, funded by the European Research Council (ERC) under the European Union’s Horizon 2020 research and innovation programme (grant agreement No 771113). 

 \vspace{1ex}
\end{minipage}
\hspace{0.01\linewidth}
\begin{minipage}{0.05\linewidth}
 \vspace{0.05cm}
\raisebox{-0.25\height}{\includegraphics[trim =0mm 5mm 5mm 2mm,clip,scale=0.078]{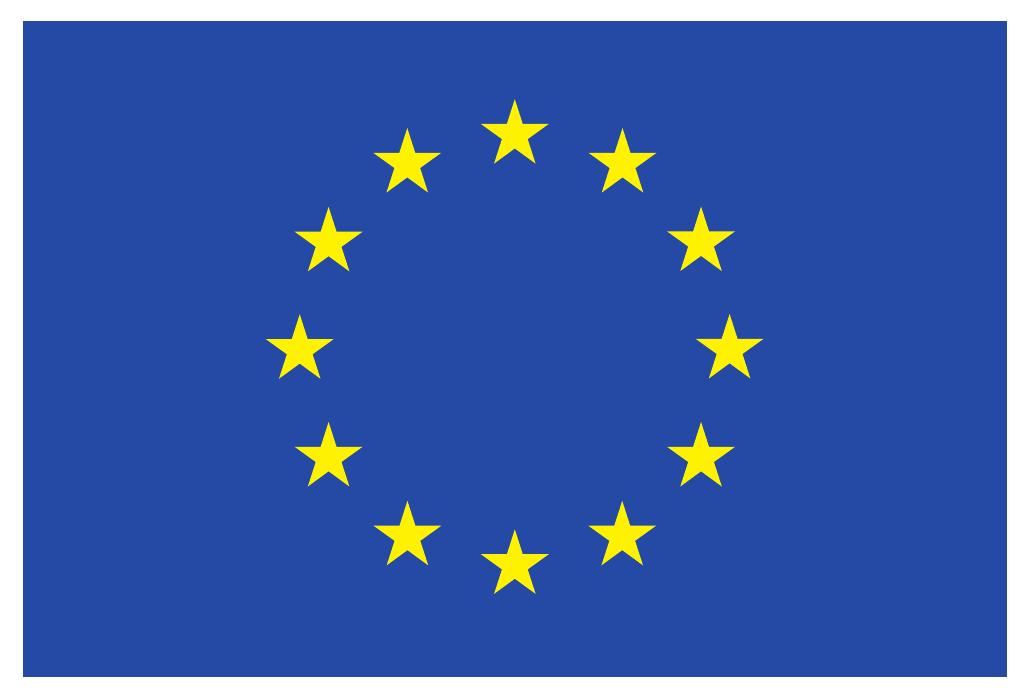}}
\vspace{0.05cm}

\end{minipage}

The first author also gratefully acknowledges the support of the Academy of Finland through project 314062 from the ICT 2023 call on Computation, Machine Learning and Artificial Intelligence.

The second author is supported by grant 2014-39 from the Swedish Research Council,
which funds the Centre for Linguistic Theory and Studies in Probability (CLASP) in the Department of
Philosophy, Linguistics, and Theory of Science at the University of Gothenburg.
\bibliography{nli}
\bibliographystyle{acl_natbib}

\end{document}